\documentclass{article}

\usepackage{arxiv}

\usepackage[utf8]{inputenc}
\usepackage[T1]{fontenc}
\usepackage{hyperref}
\usepackage{url}
\usepackage{booktabs}
\usepackage{amsfonts}
\usepackage{nicefrac}
\usepackage{microtype}
\usepackage{graphicx}
\usepackage{natbib}
\usepackage{doi}
\usepackage{amsmath,amssymb}
\usepackage{xcolor}
\usepackage{colortbl}
\usepackage{svg}

\title{\textit{OmniVLA-RL}: A Vision-Language-Action Model with Spatial Understanding and Online RL}

\author{
    Haoxiang Jie$^{1,*\dagger}$ \quad 
    Yaoyuan Yan$^{1,*}$ \quad 
    Xiangyu Wei$^{3}$ \quad 
    Kailin Wang$^{1}$ \quad 
    Hongjie Yan$^{2,4}$ \quad \\
    Zhiyou Heng$^{1}$ \quad 
    Daocheng Chen$^{1}$ \\[2ex] 
    $^{1}$AI Lab, Country Garden Services \quad 
    $^{2}$Omni AI \quad 
    $^{3}$VBot \quad 
    $^{4}$East China Normal University \\
    {\tt\small jiehaoxiang@bgyfw.com} 
    \thanks{$^*$ Equal contribution \quad $^\dagger$ Corresponding author}
}

\renewcommand{\shorttitle}{OmniVLA-RL}

\hypersetup{
    pdftitle={OmniVLA-RL: A Vision-Language-Action Model with Spatial Understanding and Online RL},
    pdfsubject={cs.RO, cs.CV, cs.LG},
    pdfauthor={Haoxiang Jie, Yaoyuan Yan, Xiangyu Wei, Kailin Wang, Hongjie Yan, Daocheng Chen},
    pdfkeywords={vision-language-action model, spatial intelligence, reinforcement learning, flow matching},
}

\begin{document}
\maketitle

\begin{abstract}
Visual-Language-Action (VLA) models represent a paradigm shift in embodied AI, yet existing frameworks often struggle with imprecise spatial perception, suboptimal multimodal fusion, and instability in reinforcement learning. To bridge these gaps, we propose OmniVLA-RL, a novel architecture that leverages a Mix-of-Transformers (MoT) design to synergistically integrate reasoning, spatial, and action experts. Furthermore, we introduce Flow-GSPO, which reformulates flow matching as a Stochastic Differential Equation (SDE) process and integrates it with Group Segmented Policy Optimization (GSPO) to enhance action precision and training robustness. Extensive evaluations on the LIBERO and LIBERO-Plus benchmarks demonstrate that OmniVLA-RL achieves decent overall performance and surpasses mainstream existing methods, effectively overcoming the fundamental limitations of current VLA models.
\end{abstract}


\section{Introduction}

Vision Language Action (VLA) models have shown significant potential for embodied AI, which equip robots with the ability to execute human language instructions in complex visual environments~\cite{Beta01}. VLA models~\cite{Beta02, Beta03} are usually improved based on the basic vision language model (VLM) and generate the robot action through a lightweight action head. General VLMs boast robust scene understanding and instruction parsing capabilities, yet they have certain limitations in spatial perception---for instance, struggling to accurately output the 3D positions and dimensions of objects. This, to a certain extent, impairs the spatial precision of VLA models. However, robotic manipulation tasks typically demand the ability for precise 3D perception of targets or the environment to achieve accurate object grasping, manipulation and obstacle avoidance. Therefore, how to effectively enhance spatial perception capabilities has become a major challenge for current VLA tasks.

Many existing VLA methods can be classified into early fusion and late fusion approaches based on the different stages at which spatial features are incorporated into the VLA pipeline. These two types of methods typically do not alter the structure of the basic VLM model, and only introduce spatial information into the feature encoder module before the VLM network or the action generation head after the VLM. For example, Evo-0~\cite{Beta04} first subjects the RGB image features, extracted by ViT encoder~\cite{Beta05}, and the spatial features, extracted by the VGGT Spatial encoder~\cite{Beta06}, to cross-attention computation in the Fusion Layer, then feeds the fused visual features and language tokens into the general VLM model. SpatialVLA~\cite{Beta07} injects 3D information into the RGB image features, extracted by SigLIP via Ego3D position encoding, and then passes the fused features into the fundamental VLM model---PaLiGemma2~\cite{Beta08}. By contrast, FALCON~\cite{Beta09} adopts a late fusion architecture. It extracts spatial features by introducing the VGGT encoder, and performs cross-attention computation with action tokens output by the standard VLA model to enhance the spatial precision of the output actions. However, the early and late fusion schemes only modify the model at the encoder and action head levels, without touching the core large model component. As a result, they cannot effectively integrate linguistic instructions, visual semantics, spatial information and robotic actions in a highly efficient manner.

On the other hand, considering that the training process of large language models can generally be divided into three stages: pre-training, supervised fine-tuning, and reinforcement learning. Recently, some researchers~\cite{Beta10, Beta11, Beta12} have attempted to introduce large model reinforcement learning methods, such as PPO~\cite{Beta13} and GRPO~\cite{Beta14}, into robotic manipulation tasks to further enhance the generalization performance of VLA models and address issues such as data dependency caused by imitation learning. However, the PPO architecture requires the design of a value model with dimensions close to those of the action model, and the entire reinforcement learning process is relatively complex, making the training of VLA models extremely cumbersome. While the GRPO-based approach eliminates the need for an additional fitted value model, flaws in the design of its token importance ratio cause instability and a high tendency to collapse in its reinforcement learning training process~\cite{Beta15}.

To address the aforementioned problems and challenges, we propose OmniVLA-RL in this paper. By integrating the spatial perception expert, action expert and pre-trained VLM via the MoT architecture, we enable the flow and interaction of linguistic instruction features, visual semantic features, and 3D spatial features within the Transformer layers of the large model, and directly generate high-precision spatial action trajectories. Meanwhile, in the reinforcement learning phase, we introduce the Flow-GSPO method. As GSPO~\cite{Beta15} is designed for sequence sampling, it is naturally more compatible with the action sequences generated by VLA models and also offers better stability compared to GRPO. Furthermore, since the OmniVLA framework generates action trajectories using Flow Matching, it adopts a deterministic denoising path following the ODE process. To effectively integrate it with the GSPO algorithm and satisfy the stochasticity requirement of reinforcement learning, we modify the Flow Matching method and convert it into an SDE process. In summary, our contributions are as follows:
\begin{itemize}
    \item We propose \textbf{OmniVLA-RL}, a unified vision-language-action framework built upon a Mixture-of-Transformers (MoT) backbone. By jointly integrating a Spatial Expert, a Reasoning Expert, and an Action Expert within shared Transformer layers, our architecture enables deep, bidirectional interaction among linguistic instruction features, visual semantic features, and 3D spatial features, overcoming the representational bottleneck of existing early- and late-fusion approaches.
    \item We introduce a novel \textbf{Block-wise Causal Attention} mechanism that explicitly decouples spatial-semantic prefix tokens from action suffix tokens. This design allows uncontaminated scene understanding while enforcing autoregressive causality during action generation, ensuring both perceptual fidelity and execution coherence.
    \item We present \textbf{Flow-GSPO}, a principled online reinforcement learning method tailored for flow-matching-based VLA models. By reformulating the deterministic ODE denoising process as a Stochastic Differential Equation (SDE) via the Fokker-Planck equation, and optimizing at the action-block level using GSPO, Flow-GSPO achieves stable stochastic exploration while avoiding the token-level bias and training instability of existing GRPO-based approaches.
    \item Extensive experiments on the \textbf{LIBERO} and \textbf{LIBERO-Plus} benchmarks demonstrate that OmniVLA-RL achieves state-of-the-art performance, attaining an average success rate of \textbf{97.6\%} on LIBERO and significantly outperforming PPO and GRPO baselines on LIBERO-Plus in both convergence speed and final performance.
\end{itemize}

\section{Related Works}

\subsection{Spatial Perception Models}

How to accurately perceive the surrounding 3D environment based on 2D images is one of the key challenges in the fields of autonomous driving and robotics~\cite{Beta16, Beta17,Beta62}. The ability to precisely identify the position, size and orientation of targets, and even obtain their motion information in 3D space, determines the success of obstacle avoidance for vehicles or mobile robots~\cite{Beta18, Beta19,Beta63} and the accuracy of target grasping for manipulation robots. In recent years, numerous innovative methods have been validated and adopted in the field of spatial perception. For example, in the autonomous driving scenario, Li et al.\ proposed the BevFormer algorithm~\cite{Beta20}, which leverages the Transformer network to convert forward-view RGB images into BEV features, enabling object detection in 3D space. Xie et al.~\cite{Beta21} and Li et al.~\cite{Beta22} proposed the M2BEV and FastBEV algorithms, which integrate the principle of camera-based ray projection into convolutional networks, achieving high-performance 3D spatial perception on low-computing-power in-vehicle platforms. In the robotics field, the VoxPoser algorithm of the Li FeiFei team~\cite{Beta23} improved the model's spatial observation capability by introducing a 3D Value Map into the VLM. Fang et al.\ further proposed the Embodied Spatial Model in the FALCON algorithm~\cite{Beta09} and injected 3D information into the action head, achieving efficient object grasping in 3D space.

\subsection{Vision-Language-Action Models}

Benefiting from the rapid development and large-scale deployment of large language models~\cite{Beta24, Beta25} and VLMs~\cite{Beta26, Beta27}, researchers have attempted to integrate large models to resolve autonomous driving and robotic manipulation problems in the end-to-end manner, and have proposed a series of Vision-Language-Action (VLA) models~\cite{Beta28, Beta29}. For example, recent approaches such as RT-2~\cite{Beta30} and OpenVLA~\cite{Beta02} build on pre-trained VLMs and integrate an autoregressive prediction head to output robotic motion trajectories. Specifically, the PI series of algorithms~\cite{Beta31, Beta32, Beta12} introduce an action chunking architecture based on flow matching and achieve promising performance. Shukor et al.~\cite{Beta01} proposed a tiny and efficient VLA method based on the Smol-vlm~\cite{Beta33} algorithm, enabling robotic manipulation tasks on consumer-grade GPU. Furthermore, VLA-RL~\cite{Beta10} and PI-RL~\cite{Beta11} integrate reinforcement learning with VLA models to enhance the generalization ability and success rate.

\subsection{Generative Models in Embodied Intelligence}

Recently, generative large models have achieved significant breakthroughs in the fields of image generation, video generation and speech generation. Diffusion models, consistency models and flow matching constitute the three core paradigms of generative large models. The Denoising Diffusion Probabilistic Models (DDPM) series of algorithms~\cite{Beta34, Beta35} gradually transform the data distribution into a simple prior distribution (e.g., the standard Gaussian distribution) through a discrete noise-adding process, and then generate new data from the prior distribution by learning the reverse denoising process, thus completing the modeling of high-dimensional manifold distributions. Optimization-based diffusion methods represented by score-based generative model (SGM)~\cite{Beta36} and its variants~\cite{Beta37, Beta38} learn the noise network by minimizing the noise loss function. Consistency Models~\cite{Beta39, Beta40} take pre-trained diffusion models as the teacher model, and compress their generation capability for multi-step inference into the student model via consistency distillation. These methods enable the student model to directly generate high-quality samples from noise in a single step or a few steps, thus addressing the bottleneck of slow inference in traditional diffusion models. Flow Matching (FM)~\cite{Beta41, Beta42} is a generative modeling framework based on Continuous Normalizing Flows (CNFs), whose core lies in learning time-dependent vector fields. It continuously transforms a simple initial distribution (e.g., Gaussian noise) into the target data distribution through Ordinary Differential Equations (ODEs), converting generative modeling into vector field regression. FM eliminates the need for complex likelihood calculation or multi-step iteration, and features both high training efficiency and stable generation performance.

In essence, the above methods realize the probabilistic distribution modeling of data samples in the manifold space, and are not limited to the specific modal characteristics of data (e.g., images and speech) in principle. Considering that both images and robotic actions can be regarded as ``distributions on high-dimensional manifolds'', some scholars have attempted to transfer generative algorithms from the image field into the robotics field for action generation. Representative algorithms include Diffusion-policy~\cite{Beta43}, Consistent-policy~\cite{Beta44}, CEED-VLA~\cite{Beta45} and Flow-policy~\cite{Beta46}.

\subsection{Reinforcement Learning for Large Models}

The core objective of reinforcement learning is to optimize large models through interactive feedback, making their behavioral outputs more aligned with human preferences, task requirements, or environmental constraints, which complements the imitation training in the supervised fine-tuning phase. Among them, the policy gradient method~\cite{Beta47} is a classic reinforcement learning framework that calculates an unbiased estimate of the policy gradient and optimizes the policy via gradient ascent to maximize cumulative reward. Trust Region Policy Optimization (TRPO)~\cite{Beta48} introduces an advantage function to estimate the action's reward value and constrains the divergence between the old and new policies using the Kullback-Leibler (KL) divergence, thus preventing training collapse caused by an excessively large policy update step size. Proximal Policy Optimization (PPO)~\cite{Beta13} is one of the main reinforcement learning methods for large models. Based on TRPO, it simplifies the constrained optimization problem by converting the KL divergence constraint into a clipping term in the objective function. Group Relative Policy Optimization (GRPO)~\cite{Beta14} is an improved PPO-based algorithm proposed in DeepSeek-Math; it addresses the issue that estimating advantage values requires the simultaneous use of both a reward model and a value model, and replaces the traditional value model with group relative advantage estimation.

\section{Preliminaries}

In this section, we briefly introduce the GSPO algorithm and the flow matching algorithm, and elaborate on their basic principles to facilitate the discussion in Section~\ref{sec:method}.

\subsection{GSPO}

The goal of RL is to optimize the parameters of policy $\pi_\theta$ so as to maximize the expected cumulative reward under the horizon of $H$:
\begin{align}
  \mathcal{J}_{RL}(\theta) = \mathbb{E}_{\pi_\theta}\left[\sum_{t=0}^H \gamma^t R(s_t, a_t)\right]
\end{align}
where $\gamma$ is the discount factor, $R$ is the reward function, and $s_t$, $a_t$ denote the state and action at time $t$, respectively.

Furthermore, GSPO defines the importance ratio based on sequential likelihood, and its optimization objective can be described as follows:
\begin{equation}
\begin{split}
\mathcal{J}_{GSPO}(\theta) = \mathbb{E}_{x \sim \mathcal{D}, \{y_i\}_{i=1}^G \sim \pi_{\theta_{\text{old}}}} \bigg[ & \frac{1}{G} \sum_{i=1}^{G} \min \big( s_i(\theta)\widehat{\textbf{A}}_i, \\
& \text{clip}\big(s_i(\theta), 1-\varepsilon, 1+\varepsilon\big)\widehat{\textbf{A}}_i \big)\bigg]
\end{split}
\label{eq:gspo_loss}
\end{equation}
where the group-based advantage estimation is:
\begin{align}
\hat{A}_i = \frac{r(x,y_i) - \text{mean}\left(\{r(x,y_j)\}_{j=1}^G\right)}{\text{std}\left(\{r(x,y_j)\}_{j=1}^G\right)}
\end{align}
and the importance ratio is:
\begin{align}
 s_i(\theta) =\exp\left( \frac{1}{|y_i|} \sum_{t=1}^{|y_i|} \log \frac{\pi_\theta(y_{i,t}|x, y_{i,<t})}{\pi_{\theta_{\text{old}}}(y_{i,t}|x, y_{i,<t})} \right)
\end{align}

\subsection{Flow Matching}

The core of flow matching models lies in learning a continuous normalizing flow that maps a simple initial distribution (e.g., the Gaussian distribution) to the target data distribution. In mathematical terms, a flow refers to a continuous process of the temporal evolution of data samples, whose evolution rate and direction can typically be described by the following ODE:
\begin{align}
\frac{\mathrm{d}\mathbf{x}_t}{\mathrm{d}t} = \mathbf{v}_t(\mathbf{x}_t, t)
\end{align}
where $\mathbf{v}_t(\cdot)$ is the velocity field.

Assume that $x_0 \sim p_0(x) = \mathcal{N}(x \mid 0, I)$ denotes the initial sampling, and $x_1 \sim p_1(x)$ is the sampling from the true data distribution. For an arbitrary time step $t \sim U(0, 1)$, we have $x \sim p_t(x)$. In the ReFlow algorithm, the probability density path can be obtained via linear interpolation, i.e., $x = (1-t)x_0 + tx_1$. The flow matching model regresses the target velocity field $\boldsymbol{\mu}(x)$ using $\mathbf{v}_\theta(x, t)$ through a neural network and optimizes the model parameter $\theta$ by minimizing the following objective:
\begin{align}
\mathcal{L}_{\text{FM}}(\theta) = \mathbb{E}_{t, p_t(x)} \left\| \boldsymbol{v}_\theta(x) - \boldsymbol{u}(x) \right\|^2
\end{align}

Due to $\boldsymbol{\mu}(x)$ being difficult to obtain directly in practical scenarios, Lipman et al.\ further introduce the conditional velocity field $\boldsymbol{u}(x|x_1)$ and derive the Conditional Flow Matching (CFM) model:
\begin{align}
\mathcal{L}_{\text{CFM}}(\theta) = \mathbb{E}_{t, q(x_1), p_t(x|x_1)} \left\| \boldsymbol{v}_\theta(x) - \boldsymbol{u}(x|x_1) \right\|^2
\end{align}

\section{Methodology}
\label{sec:method}

In this section, we elaborate on OmniVLA-RL, a vision-language-action model with spatial understanding and online RL. In Section~\ref{sec:problem}, we present the problem formulation and definition of basic notations. We then describe the overall architecture in Section~\ref{sec:arch}, and detail the unified spatial-reasoning-action model in Section~\ref{sec:mot}. In Section~\ref{sec:flowgspo}, we elaborate on the online reinforcement learning process that integrates GSPO with Flow Matching.

\subsection{Problem Definition and Notation}
\label{sec:problem}

To integrate online RL with VLA tasks, we first model the robotic manipulation task as a Markov process, denoted by the tuple $M=(\mathcal{S},\mathcal{A},\mathcal{P},\mathcal{R},\rho_0)$. Here, the robot state space $\mathcal{S}$ consists of RGB images $I$, linguistic instructions $L$ and proprioceptive states $\mathcal{S}_{\text{prop}}$; $\mathcal{A}$ denotes the action space; $\mathcal{P}(s_{t+1}|s_t,a_t)$ is the state transition function; $\mathcal{R}(s_t,a_t)$ represents the reward function; and $\rho_0$ is the initial state distribution. At timestep $t$, the robot observation is defined as $o_t\triangleq s_t$, and the agent samples an action $a_t\sim\pi_\theta(\cdot|s_t)\in\mathcal{A}$ based on the current observation, where $\pi_\theta(\cdot|s_t)$ is the VLA model and $\theta$ denotes the neural network model parameters.

\subsection{Architecture Overview}
\label{sec:arch}

\begin{figure}[t]
\centering
\includegraphics[width=\textwidth]{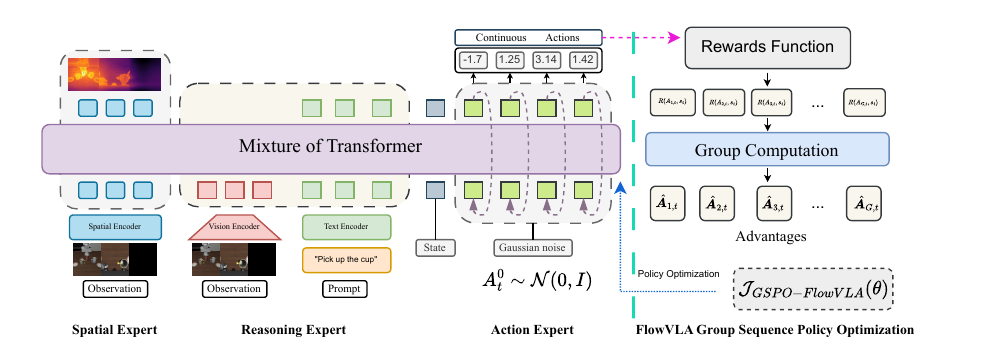}
\caption{Overall architecture of OmniVLA-RL. The VLA model adopts a Mixture-of-Transformers (MoT) backbone shared across three experts. The \textit{Reasoning Expert} (centre) encodes multi-view RGB observations via a Vision Encoder and task instructions via a Text Encoder, producing semantic and linguistic tokens. The \textit{Spatial Expert} (left) extracts fine-grained 3D structural features from multi-view scenes using a Spatial Encoder, and a lightweight Spatial Decoder is appended as an auxiliary supervision head during training. The \textit{Action Expert} (right) generates action trajectories autoregressively conditioned on the fused spatial-semantic representations. All three experts share the same Transformer layers and interact via the proposed Block-wise Causal Attention, which treats spatial and semantic tokens as an omni-visible prefix while enforcing causal constraints on action tokens. The online RL module (Flow-GSPO) fine-tunes the entire model by optimising action-block-level policy rewards through stochastic flow matching.}
\label{fig:architecture}
\end{figure}

As illustrated in Figure~\ref{fig:architecture}, OmniVLA-RL is composed of a VLA model and an online reinforcement learning module. The VLA model adopts the MoT architecture, which consists of a Spatial Expert, a Reasoning Expert, and an Action Expert. Based on the fundamental VLM, the Reasoning Expert takes multi-view observations $I_t$ and instructions $L$ as input to extract linguistic and visual information in the scene. The Spatial Expert is mainly responsible for extracting spatial features from multi-view scenes, and performs attention computation with the transmitted linguistic and visual features within the Transformer network to obtain spatial features associated with the linguistic instruction $L$. With the spatial representations, semantic features transmitted from the first two experts and linguistic supervision as inputs, the Action Expert maps them end-to-end to executable robotic actions.

\subsection{Unified Spatial-Reasoning-Action Model}
\label{sec:mot}

OmniVLA-RL adopts a Mixture-of-Transformers (MoT) backbone~\cite{Beta49}, featuring a novel tri-expert mechanism designed to disentangle and optimize multifaceted representations. The Reasoning Expert leverages a VLM to process multi-view observations $O$ and linguistic instructions $L$ to extract high-level semantic embeddings and visual priors. The Spatial Expert is dedicated to capturing granular spatial representations across multi-view scenes; by performing attention mechanisms within the Transformer blocks, it integrates the transferred semantic information to yield task-relevant spatial-semantic features. The Action Expert takes the fused spatial-semantic representations and linguistic supervision as input, facilitating an end-to-end mapping from multimodal perceptions to executable robotic control signals. This hierarchical expert design enables OmniVLA-RL to bridge the gap between abstract reasoning and precise spatial groundedness.

\subsubsection{Reasoning Expert.}
To endow the robot with robust instruction-following and scene-understanding capabilities, the \textbf{Reasoning Expert} is initialized with a pre-trained VLM optimized on large-scale image-text datasets, ensuring the integration of extensive commonsense priors. At each temporal step, the module employs \textbf{SigLIP}~\cite{Beta58} as the vision encoder to extract high-level semantic features $z_{sem} \in \mathbb{R}^{n \times d}$ from the multi-view observations $\mathcal{O}=\{O_i\}_{i=1}^M$. These visual embeddings are concatenated with linguistic tokens $z_{lang}$ and fed into a \textbf{decoder-only Transformer}~\cite{Beta59} backbone to facilitate cross-modal alignment by modeling the conditional distribution $p(z_{lang} \mid z_{sem})$. This architectural design empowers the Reasoning Expert to perform sophisticated semantic reasoning, where the resulting latent variables serve as a global context to provide a foundational representational scaffold for subsequent modules.

\subsubsection{Spatial Expert.}
The \textbf{Spatial Expert} is designed to extract comprehensive structural information from multi-view observations $\mathcal{O} = \{O_i\}_{i=1}^M$, providing essential geometric grounding for action generation. Fine-grained manipulation tasks necessitate an intricate understanding of physical spatial relations and object configurations. However, conventional high-level VLMs frequently suffer from a loss of fine-grained spatial attributes.

To mitigate this, we employ \textbf{VGGT}~\cite{Beta60} to extract granular features, which are subsequently integrated into the Transformer backbone to yield the spatial representation $z_{spatial}$. To facilitate optimization, a lightweight Transformer decoder is appended to the final hidden state $h_i \in \mathbb{R}^{C \times d}$ as a \textbf{spatial auxiliary head}. This head is supervised via spatial-centric pretext tasks during training to distill geometric knowledge. Notably, this auxiliary component is decoupled from the downstream inference pipeline and does not participate in the final action generation.

\subsubsection{Action Expert.}
The \textbf{Action Expert} is responsible for generating precise control commands conditioned on multimodal observations and linguistic instructions. By explicitly incorporating spatial priors, this module synergizes high-level semantic observations with fine-grained spatial features, thereby enforcing spatial consistency and physical feasibility in action synthesis.

Implementation-wise, we adopt an \textbf{action chunking} strategy, where action sequences are mapped into the Transformer's latent space via a linear projector. To achieve high-fidelity trajectory generation, we utilize \textbf{Conditional Flow Matching (CFM)}~\cite{Beta51} to model the action distribution conditioned on the joint representation of spatial attributes, semantic features, and linguistic directives:
\begin{equation}
    a_t \sim p(a \mid z_{spatial}, z_{sem}, z_{lang})
\end{equation}
This framework ensures both the precision of action generation and the computational efficiency required for real-time robotic execution.

\subsubsection{Block-wise Causal Attention.}
To integrate heterogeneous multimodal representations within a unified Transformer architecture, we propose a \textbf{Block-wise Causal Attention} mechanism that explicitly decouples and fuses information across modalities through a meticulously designed mask matrix.

Specifically, the tokens from both the Reasoning and Spatial Experts are treated as an \textbf{omni-visible prefix}, facilitating bidirectional cross-modal alignment between granular spatial patches and macro-semantic contexts under the global guidance of task prompts. This ensures the construction of a physically grounded environment representation prior to decision-making. For the \textbf{Action Suffix}, we enforce strict causal and unidirectional constraints. While action chunks maintain access to the complete prefix information flow, they adhere to autoregressive causality internally. Crucially, the prefix modules are restricted from attending to the latent noise within subsequent action blocks, preventing stochastic noise from the diffusion sampling process from contaminating the scene understanding.

\begin{figure}[t]
\centering
\includegraphics[width=0.48\textwidth]{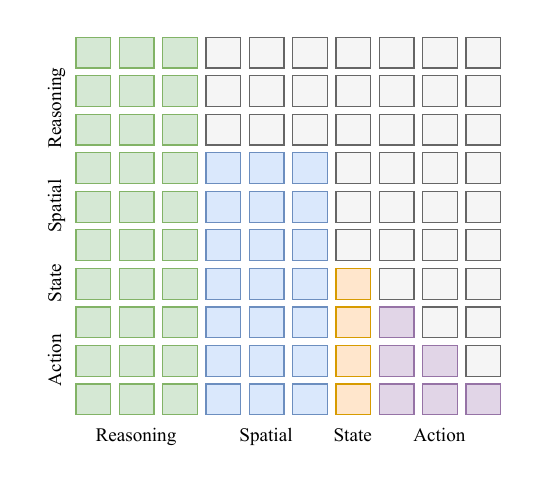}
\caption{Block-wise Causal Attention mask of OmniVLA-RL. Tokens from the Spatial and Reasoning Experts form an \textit{omni-visible prefix} with full bidirectional attention among themselves (dark blocks), enabling rich cross-modal alignment between spatial and semantic representations. Action tokens form a \textit{causal suffix}: each action chunk can attend to the entire prefix but is restricted to attending only to preceding action tokens within the suffix (lower-triangular pattern). Prefix tokens are blocked from attending to action tokens (white blocks), preventing stochastic denoising noise from contaminating scene understanding.}
\label{fig:attention}
\end{figure}

\subsection{Flow-GSPO}
\label{sec:flowgspo}

\subsubsection{Stochastic Flow Matching.}
Let the continuous action sequence generated by conditional flow matching be $A_t=[a_{t,0},\dots,a_{t,H-1}]$, the size of the iterative steps is $\delta$, $K$ denotes the denoising steps with $\delta=\frac{1}{K}$. In addition, $A_t^\tau$ represents the denoising action in the $\tau$-th step, and $A_t^0\sim\mathcal{N}(0,I)$. We further adopt the conditional probability $p(A_t^\tau|A_t)$ from the Rectified Flow framework and set the ground-truth vector field $\boldsymbol{u}(A_t^\tau|A_t) = A_t - \varepsilon$ (random noise $\varepsilon\sim\mathcal{N}(0,I)$), so we obtain the following equation:
\begin{align}
A_t^{\tau+\delta} = A_t^\tau + \delta\,\boldsymbol{v}_\theta(A_t^\tau, s_t)
\end{align}
In reinforcement learning, the exploration of actions is required to be stochastic. To this end, we refer to~\cite{Beta11, Beta10} and transform the ODE into the following SDE via the Fokker-Planck equation, where the injection of random noise renders the action generation process differentiable and amenable to stochastic exploration:
\begin{align}
\mathrm{d}A_t^\tau = \left[\boldsymbol{v}_\theta(A_t^\tau, s_t) + \frac{\sigma_\tau^2}{2} \left( A_t^\tau + (1 - \tau) \boldsymbol{v}_\theta(A_t^\tau, s_t) \right) \right] \mathrm{d}\tau + \sigma_\tau \mathrm{d}w_\tau
\end{align}
We discretize the above equation using the Euler-Maruyama method, yielding the update formula:
\begin{align}
A_t^{\tau+\delta} = A_t^\tau + \left[\boldsymbol{v}_\theta(A_t^\tau, s_t) + \frac{\sigma_\tau^2}{2} \left( A_t^\tau + (1 - \tau) \boldsymbol{v}_\theta(A_t^\tau, s_t) \right) \right] \delta + \sigma_\tau \sqrt{\delta}\, \epsilon
\end{align}
According to the above equation, we can derive that the transition probability $p\bigl(A_t^{\tau+\delta} \mid A_t^\tau, s_t\bigr) \sim \mathcal{N}\bigl(\mu_\tau, \Sigma_\tau\bigr)$ is an isotropic Gaussian distribution, where $\mu_{\tau}$ and $\Sigma_{\tau}$ are:
\begin{align}
\mu_{\tau} = A_{t}^\tau + \left[ v_\theta\bigl(A_{t}^\tau, s_t\bigr) + \frac{\sigma_\tau^2}{2} \bigl( A_{t}^\tau + (1 - \tau) v_\theta\bigl(A_{t}^\tau, s_t\bigr) \bigr) \right] \delta
\end{align}
\begin{align}
\Sigma_{\tau} = \sigma_\tau^2 \delta I
\end{align}

\subsubsection{GSPO on Stochastic Flow Matching.}
To avoid the problems of single-step bias accumulation and disruption of action continuity caused by token-based optimization in algorithms such as GRPO, we take the action block $A_{t}$ generated by the VLA task as a sequence-level optimization unit and integrate it into the online reinforcement learning framework of GSPO. Assuming the group size is $G$, for each state $s_t$, the agent samples and generates $G$ action sequences $\{A_{t,i}\}_{i=1}^G$, and the likelihood of each action sequence is given by:
\begin{equation}
\pi_\theta(A_{t,i} \mid s_t) = \prod_{\tau=0}^{K-1} p_\theta\bigl(A_{t,i}^{\tau+\delta} \mid A_{t,i}^\tau, s_t\bigr) = \prod_{\tau=0}^{K-1} \mathcal{N}\bigl(A_{t,i}^{\tau+\delta} \mid \mu_{\tau,i}, \Sigma_{\tau,i}\bigr)
\end{equation}
Let $|A_{t,i}| = H \times K$ (action block length $\times$ denoising steps). The action block-level importance ratio is:
\begin{align}
s_{i,t}(\theta) = \left( \frac{\pi_\theta(A_{i,t}|s_t)}{\pi_{\theta_{\text{old}}}(A_{i,t}|s_t)} \right)^{\frac{1}{|A_{i,t}|}} = \exp\left( \frac{1}{|A_{i,t}|} \sum_{\tau=0}^{K-1} \log \frac{p_\theta(A_{t,i}^{\tau+\delta}|A_{t,i}^\tau, s_t)}{p_{\theta_{\text{old}}}(A_{t,i}^{\tau+\delta}|A_{t,i}^\tau, s_t)} \right)
\end{align}
For the sequences of $G$ action blocks of each state $s_t$, we calculate the advantage of the action block using the normalization of task rewards:
\begin{align}
\hat{\boldsymbol{A}}_{i,t} = \frac{R_{\text{total}}(A_{i,t}, s_t) - \text{mean}\left(\{R_{\text{total}}(A_{j,t}, s_t)\}_{j=1}^G\right)}{\text{std}\left(\{R_{\text{total}}(A_{j,t}, s_t)\}_{j=1}^G\right)}
\end{align}
where $R_{\text{total}}(A_{i,t}, s_t) = \sum_{h=0}^{H-1} \gamma^h R(s_t, a_{t,i,h})$ denotes the cumulative reward of the action block sequence.

To further enhance the stability of training, we additionally introduce the KL divergence between the old and new policies to avoid drastic changes during policy iteration, yielding the final optimization objective:
\begin{equation}
\begin{aligned}
\mathcal{J}_{\text{Flow-GSPO}}(\theta)
&= \mathbb{E}_{s_t \sim D, \{A_{i,t}\}_{i=1}^G \sim \pi_{\theta_{\text{old}}}(\cdot|s_t)}
\Bigg[ \frac{1}{G} \sum_{i=1}^G \min\Big( s_i(\theta) \hat{\boldsymbol{A}}_{i,t}, \\
& \qquad \text{clip}\big( s_i(\theta), 1 - \varepsilon, 1 + \varepsilon \big) \hat{\boldsymbol{A}}_{i,t} \Big)
- \beta D_{\text{KL}}\big(\pi_\theta \,\|\, \pi_{\text{old}}\big)
\Bigg]
\end{aligned}
\end{equation}
Here, $D_{\text{KL}}$ denotes the action block-level KL divergence between the old and new policies:
\begin{equation}
D_{\text{KL}}(\pi_\theta \parallel \pi_{\text{old}}) = \mathbb{E}_{s_t, A_t \sim \pi_{\text{old}}}\left[ \log \frac{\pi_{\theta}(A_t|s_t)}{\pi_{\text{old}}(A_t|s_t)} \right]
\end{equation}

\subsubsection{Gradient Analysis.}
Based on the objective function of Flow-GSPO, we can derive its gradient with respect to the parameter $\theta$. Ignoring the clipping term, the gradient of the main term is:
\begin{equation}
\begin{aligned}
\nabla \mathcal{J}_{\text{main}}(\theta)
= \mathbb{E}_{s_t \sim D, \{A_{i,t}\}_{i=1}^G \sim \pi_{\theta_{\text{old}}}(\cdot|s_t)}
\Bigg[ \frac{1}{G} \sum_{i=1}^G s_{t,i}(\theta) \hat{\boldsymbol{A}}_{i,t} \, \nabla_\theta \log s_{t,i}(\theta) \Bigg]
\end{aligned}
\end{equation}
Combining the expressions for $\log s_{t,i}(\theta)$ and $\nabla_\theta \log s_{t,i}(\theta)$, we can derive:
\begin{equation}
\begin{aligned}
\nabla_\theta \mathcal{J}_{\text{main}}(\theta)
&= \mathbb{E}_{s_t \sim D, \{A_{t,i}\}_{i=1}^G \sim \pi_{\theta_{\text{old}}}(\cdot|s_t)}
\Bigg[ \frac{1}{G} \sum_{i=1}^G s_{t,i}(\theta) \hat{\textbf{A}}_{t,i}
\cdot \frac{1}{|A_{t,i}|} \sum_{\tau=0}^{K-1} \nabla_\theta \log p_\theta\bigl(A_{t,i}^{\tau+\delta} \mid A_{t,i}^\tau, s_t\bigr)
\Bigg]
\end{aligned}
\end{equation}
Since $p_\theta(\cdot)$ follows a Gaussian distribution, the gradient of its log-likelihood is:
\begin{equation}
\nabla_\theta \log p_\theta\bigl(A_{t,i}^{\tau+\delta} \mid A_{t,i}^\tau, s_t\bigr)
= \Sigma_{\tau,i}^{-1} \bigl(A_{t,i}^{\tau+\delta} - \mu_{\tau,i}\bigr) \cdot \nabla_\theta \mu_{\tau,i}
\end{equation}
Decomposing $\mu_{\tau,i}$ into a $\theta$-independent term $C_0$ and a $\theta$-dependent term $C_r$:
\begin{equation}
\mu_{\tau,i}
= \underbrace{A_{t,i}^\tau \left(1 + \frac{\sigma_\tau^2 \delta}{2}\right)}_{C_0}
+ \underbrace{\left[1 + \frac{\sigma_\tau^2(1-\tau)}{2}\right] \delta \cdot \mathbf{v}_\theta(A_{t,i}^\tau, s_t)}_{C_r}
\end{equation}
\begin{equation}
\nabla_\theta \mu_{\tau,i} = C_r \cdot \nabla_\theta \mathbf{v}_\theta\bigl(A_{t,i}^\tau, s_t\bigr)
\end{equation}
Finally, the overall gradient of Flow-GSPO is:
\begin{equation}
\begin{aligned}
\nabla_\theta \mathcal{J}_{\text{Flow-GSPO}}(\theta)
&= \mathbb{E}_{s_t \sim D, \{A_{t,i}\}_{i=1}^G \sim \pi_{\theta_{\text{old}}}(\cdot|s_t)}
\Bigg[ \frac{1}{G} \sum_{i=1}^G \left( \frac{s_{t,i}(\theta) \hat{\textbf{A}}_{t,i}}{|A_{t,i}|} + \beta \right) \\
&\quad \cdot \sum_{\tau=0}^{K-1} \Bigl( \Sigma_{\tau,i}^{-1} \bigl(A_{t,i}^{\tau+\delta} - \mu_{\tau,i}\bigr)
\bigl[1 + \frac{\sigma_\tau^2(1-\tau)}{2}\,\delta\bigr] \Bigr)
\cdot \nabla_\theta \mathbf{v}_\theta\bigl(A_{t,i}^\tau, s_t\bigr)
\Bigg]
\end{aligned}
\end{equation}

\section{Experiments}

To evaluate the efficacy of our proposed framework, we design a \textbf{three-stage training paradigm} following a progressive evolution from spatial-aware alignment to action generation, ensuring the model's ability to synergistically process visual semantics, geometric spatial cues, and execution tasks.

\begin{figure}[t]
\centering
\includegraphics[width=\textwidth]{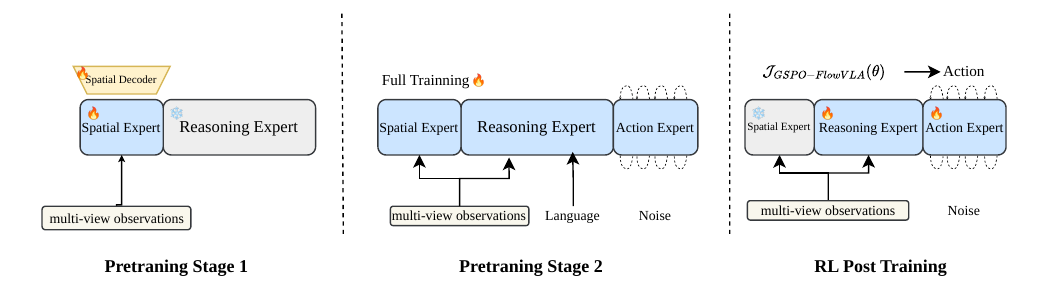}
\caption{Three-stage progressive training paradigm of OmniVLA-RL. \textbf{Stage I} (Spatial Pre-training): the Reasoning and Spatial Experts are jointly trained on large-scale 3D datasets with the Action Expert frozen, establishing stable spatial-semantic representations via point cloud, camera, and surface normal reconstruction losses. \textbf{Stage II} (Action Pre-training): the Action Expert is unfrozen and trained end-to-end on the DROID dataset using Conditional Flow Matching, bridging scene understanding with policy synthesis. \textbf{Stage III} (Online RL): the full model is fine-tuned via Flow-GSPO on task-specific environments, refining the policy through stochastic flow matching and action-block-level reward optimisation.}
\label{fig:train_stage}
\end{figure}

\subsection{Experimental Setup}

\subsubsection*{Stage I: Multimodal Spatial Perception Pre-training}
In the initial pre-training phase, our primary objective is to cultivate stable and discriminative multimodal perceptual and spatial representations. We initialize the VLM with pre-trained \textbf{PaLiGemma}~\cite{Beta52} weights, while the Spatial Expert is initialized stochastically. Both modules undergo end-to-end joint optimization within our unified framework.

Training during this stage leverages large-scale 3D datasets, with supervisory signals concentrated on structural modeling and geometric relationship reasoning. To prevent the perceptual features from developing an action bias prior to convergence, the \textbf{Action Expert remains frozen}. Inspired by \textbf{PI-3}~\cite{Beta53}, we formulate the spatial loss $\mathcal{L}_{Spatial}$ as a reconstruction-based objective:
\begin{equation}
    \mathcal{L}_{Spatial} = \mathcal{L}_{points} + \lambda_{cam} \mathcal{L}_{cam} + \lambda_{normal} \mathcal{L}_{normal}
\end{equation}
where $\mathcal{L}_{points}$, $\mathcal{L}_{cam}$, and $\mathcal{L}_{normal}$ represent the reconstruction losses for point clouds, camera parameters, and surface normals, respectively, with $\lambda$ denoting the pre-defined hyper-parameters for task balancing.

\subsubsection*{Stage II: Action Generation Pre-training}
Building upon the spatial perceptual foundation established in Stage I, the second stage focuses on scaling the model's action modeling capabilities. During this phase, we unfreeze the \textbf{Action Expert} for parameter optimization while deactivating the \textbf{Spatial Head}.

Training is conducted on the full \textbf{DROID}~\cite{Beta49} dataset. We employ the Conditional Flow Matching (CFM) loss~\cite{Beta54, Beta55} as the primary optimization objective:
\begin{equation}
    \mathcal{L}_{CFM} = \mathbb{E}_{t \sim \mathcal{U}(0,1), \mathbf{x}_0 \sim p_0} \left[ \left\| \mathbf{v}_t(\mathbf{x}_t, t; \mathbf{c}) - (\mathbf{x}_1 - \mathbf{x}_0) \right\|_2^2 \right]
\end{equation}
where $\mathbf{c}$ denotes the multimodal conditional context and $\mathbf{x}_t = t\mathbf{x}_1 + (1-t)\mathbf{x}_0$ is the interpolated action state.

\subsubsection*{Stage III: Online Reinforcement Learning with Flow-GSPO}
The third stage applies online RL to further refine the policy learned in Stage II. The model is initialised from the Stage II checkpoint with all parameters unfrozen. For each training episode, the agent generates $G$ candidate action blocks $\{A_{t,i}\}_{i=1}^{G}$ via the stochastic flow matching process, executes each block, and observes the resulting task reward. The reward signal is composed of a binary task-completion reward and a continuous gripper-alignment reward measuring the distance between the end-effector and the target object.

The policy is updated using the Flow-GSPO objective with group size $G{=}8$, clipping coefficient $\varepsilon{=}0.2$, and KL penalty weight $\beta{=}0.01$. The noise schedule follows $\sigma_\tau = \sigma_{\max}(1-\tau)$ with $\sigma_{\max}{=}0.1$. We use $K{=}10$ denoising steps per action block and an action horizon of $H{=}16$. The model is optimised with AdamW (lr $= 1{\times}10^{-5}$, weight decay $= 0.01$) for 200 RL update steps, with a rollout buffer refreshed every 10 steps.

\subsection{Simulation Benchmark Experiments}

We systematically evaluate our proposed strategy on two robotic manipulation benchmarks with fundamentally different difficulty profiles: LIBERO~\cite{Beta56} and LIBERO-Plus~\cite{Beta57}.

\textbf{LIBERO} (Benchmarking Knowledge Transfer in Lifelong Robot Learning) is designed to assess the transferability and adaptability of embodied agents across four task suites---LIBERO-Spatial, LIBERO-Object, LIBERO-Goal, and LIBERO-Long---probing spatial reasoning, object-centric manipulation, goal-conditioned completion, and moderate-horizon decision-making.

\textbf{LIBERO-Plus} is a substantially more challenging extension that introduces \textit{compositional long-horizon tasks} requiring the agent to execute multi-stage manipulation sequences (e.g., open a drawer, retrieve a specific object, and place it at a goal location while avoiding distractors) within a single episode. Compared to LIBERO, LIBERO-Plus tasks involve longer action horizons, denser object interactions, and stricter spatial precision requirements. The increased task complexity explains the substantially lower absolute success rates observed on LIBERO-Plus relative to LIBERO across all methods; the key metric of interest is therefore the \textit{relative improvement} brought by each algorithmic component.

\subsubsection{LIBERO Benchmark Results}

As summarized in Table~\ref{table:libero_results}, OmniVLA-RL delivers excellent performance across all four task suites, consistently securing the top rank in every category with an overall average success rate of \textbf{97.6\%}.

\begin{itemize}
    \item \textbf{Superiority in Complex Reasoning:} In LIBERO-Spatial and LIBERO-Goal, OmniVLA-RL surpasses the strongest baseline ($\pi_{0.5}$) by 0.4\% and 0.5\%, respectively. These gains underscore the efficacy of our tri-expert architecture in modulating spatial perception and goal-directed reasoning.
    \item \textbf{Robustness in Long-Horizon Tasks:} OmniVLA-RL achieves a \textbf{93.5\%} success rate in LIBERO-Long, a 1.1\% margin over the runner-up that is particularly significant given that long-horizon tasks are highly sensitive to compounding errors.
    \item \textbf{Performance Gains over Open-source Baselines:} Compared to $\pi_0$, OmniVLA-RL provides a substantial \textbf{21.1\%} absolute improvement in average success rate, highlighting the representational power of our three-expert architecture refined through supervised fine-tuning.
\end{itemize}

\begin{table}[t]
\centering
\caption{Performance comparison on the LIBERO benchmark. Bold indicates best performance.}
\label{table:libero_results}
\setlength{\tabcolsep}{3pt}
\resizebox{\linewidth}{!}{
\begin{tabular}{lcccccccccc}
\toprule
Method & Spatial & Rank & Object & Rank & Goal & Rank & Long & Rank & Avg & Rank \\
\midrule
Diffusion Policy & 78.5\% & 7 & 87.5\% & 7 & 73.5\% & 7 & 64.8\% & 4 & 76.1\% & 7 \\
Octo             & 78.9\% & 6 & 85.7\% & 8 & 75.1\% & 6 & 54.1\% & 7 & 74.8\% & 8 \\
OpenVLA          & 84.7\% & 5 & 88.4\% & 6 & 79.2\% & 4 & 63.7\% & 5 & 76.5\% & 6 \\
SpatialVLA       & 88.2\% & 4 & 89.9\% & 5 & 78.6\% & 5 & 55.5\% & 6 & 78.1\% & 5 \\
CoT-VLA          & 87.5\% & 3 & 91.6\% & 4 & 87.6\% & 3 & 69.0\% & 3 & 83.9\% & 4 \\
$\pi_0$          & 96.8\% & 2 & 98.8\% & 2 & 95.8\% & 2 & 85.2\% & 2 & 94.2\% & 3 \\
$\pi_{0.5}$      & 98.8\% & 1 & 98.2\% & 3 & 98.0\% & 1 & 92.4\% & 1 & 96.9\% & 2 \\
$F_1$~\cite{Beta61} & 98.2\% & 2 & 97.8\% & 4 & 95.4\% & 3 & 91.3\% & 2 & 95.7\% & 3 \\
\rowcolor{green!15} \textbf{OmniVLA-RL (Ours)} & \textbf{99.2\%} & \textbf{1} & \textbf{99.2\%} & \textbf{1} & \textbf{98.5\%} & \textbf{1} & \textbf{93.5\%} & \textbf{1} & \textbf{97.6\%} & \textbf{1} \\
\bottomrule
\end{tabular}
}
\end{table}

\subsubsection{LIBERO-Plus Benchmark Results}

\begin{figure}[t]
\centering
\includegraphics[width=\textwidth]{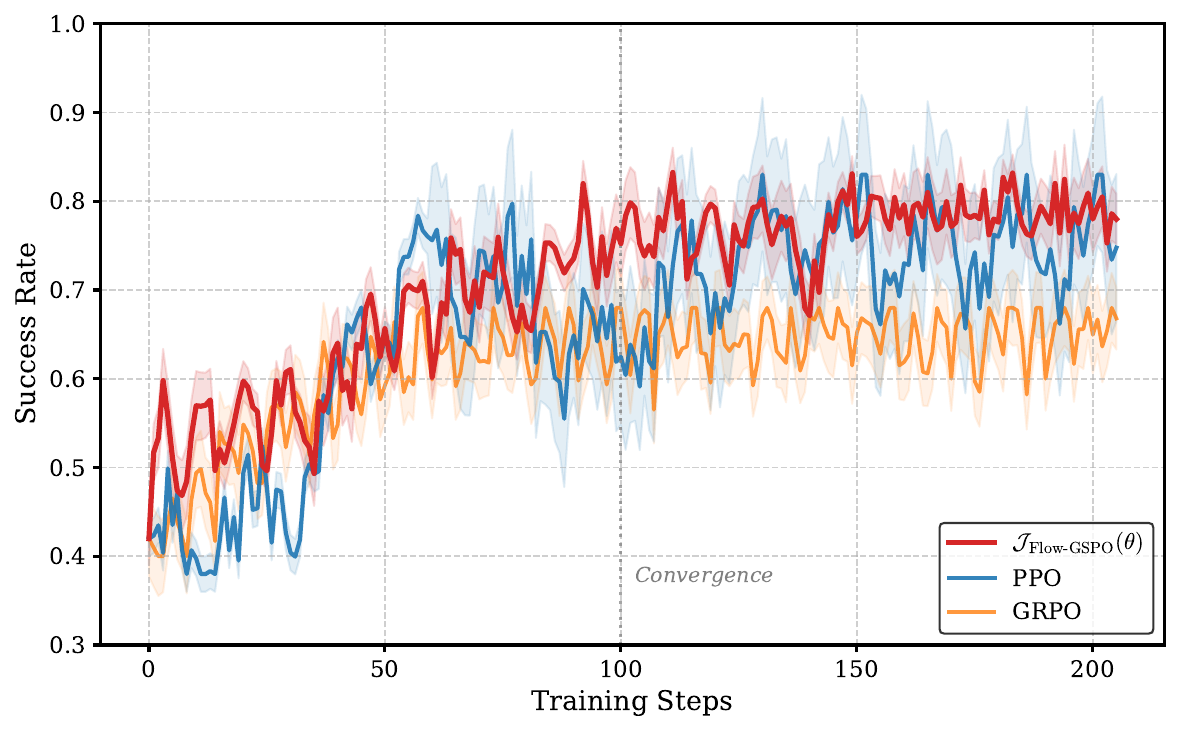}
\caption{Comparison of training success rates on the LIBERO-Plus multi-task benchmark. Flow-GSPO exhibits superior convergence speed and higher final success rates compared to PPO and GRPO baselines.}
\label{fig:libero_plus}
\end{figure}

To further investigate training dynamics and multi-task scalability, we evaluate our approach on LIBERO-Plus. Figure~\ref{fig:libero_plus} illustrates the success rate evolution of Flow-GSPO compared to PPO and GRPO baselines.

\begin{itemize}
    \item \textbf{Enhanced Sample Efficiency:} Flow-GSPO surpasses a 70\% success rate within the first 50 training steps, significantly faster than PPO and GRPO. This efficiency stems from our action block-level importance ratio, which captures sequence-level dependencies more effectively than token-based RL methods.
    \item \textbf{Superior Convergence Stability:} While PPO suffers from noticeable performance fluctuations and regressions (e.g., around step 80), Flow-GSPO maintains a consistent monotonic improvement trend, directly attributed to our segmented policy optimization and the action-block KL divergence term.
    \item \textbf{High Performance Ceiling:} In the final convergence phase (post 100 steps), Flow-GSPO maintains a success rate above \textbf{80\%}, outperforming GRPO by approximately 14.6\%. This confirms that Stochastic Flow Matching not only accelerates policy discovery but also achieves a higher performance ceiling.
\end{itemize}

\subsection{Ablation Study}

To evaluate the contribution of each core component, we conduct extensive ablation experiments on the LIBERO-Plus benchmark. As discussed above, LIBERO-Plus poses substantially greater challenges than standard LIBERO due to its compositional long-horizon task structure; accordingly, the SFT-only baseline achieves a modest 41.2\% success rate, which is consistent with the performance ceiling of imitation learning under complex multi-stage task distributions. The results are summarized in Table~\ref{table:ablation}.

\subsubsection{Superiority of the Flow-GSPO Paradigm.}
Compared to the vanilla OmniVLA-RL (SFT only), our \textbf{Flow-GSPO} achieves a \textbf{+39.1\%} absolute increase in success rate. Under identical training constraints, Flow-GSPO significantly outperforms PPO (+1.6\%) and GRPO (+14.6\%), confirming that action block-level optimization is inherently better suited for capturing the temporal dependencies and multi-modal distributions of continuous robotic trajectories than token-based policy gradient methods.

\subsubsection{Critical Impact of the Spatial Expert.}
Removing the Spatial Expert results in a \textbf{8.3\%} performance drop (from 41.2\% to 32.9\%), representing the most significant degradation among all architectural ablations. This decline underscores the necessity of spatial decoupling: without the high-resolution spatial features provided by this expert, the reasoning chain lacks the requisite precision to localize small-scale objects or handle complex occlusions.

\begin{table}[h]
\centering
\caption{Ablation study of OmniVLA-RL components on LIBERO-Plus.}
\label{table:ablation}
\begin{tabular}{lcc}
\toprule
\textbf{Configuration} & \textbf{Success Rate} & $\Delta$ \\
\midrule
OmniVLA-RL (SFT only)               & 41.2\%          & ---       \\
+ PPO                                & 78.7\%          & +37.5\%   \\
+ GRPO                               & 65.7\%          & +24.5\%   \\
+ \textbf{Flow-GSPO (Ours)}          & \textbf{80.3\%} & \textbf{+39.1\%} \\
\midrule
\rowcolor[gray]{0.9} w/o Spatial Expert (Reasoning + Action only) & 32.9\% & $-$8.3\% \\
\bottomrule
\end{tabular}
\end{table}

\section{Conclusions}

In this paper, we presented OmniVLA-RL, a robust reinforcement learning framework for embodied foundation models. By synergizing a MoT architecture with our novel Flow-GSPO optimization paradigm, OmniVLA-RL effectively addresses the challenges of complex spatial reasoning and long-horizon manipulation. The MoT structure---comprising specialized Spatial, Reasoning, and Action transformers---provides the necessary representational capacity to decouple perception and execution, while Flow-GSPO ensures stable, sample-efficient policy refinement via stochastic flow matching.

Our evaluation on the LIBERO benchmarks demonstrates that OmniVLA-RL achieves state-of-the-art results, reaching an average success rate of 97.6\% and significantly outperforming established baselines. Ablation studies highlight the Spatial Expert as a cornerstone of the MoT framework, confirming that specialized geometric grounding is essential for high-precision tasks. Furthermore, the superior convergence of Flow-GSPO over traditional RL methods validates the efficacy of action chunk-level optimization. Overall, OmniVLA-RL establishes a powerful paradigm for developing generalizable, high-performance robotic agents capable of mastering diverse and challenging operational scenarios.

\section{Limitations and Future Work}

While OmniVLA-RL achieves superior performance in benchmark evaluations, several limitations remain. First, our framework has primarily been validated within high-fidelity simulation environments; thus, the sim-to-real gap and the model's robustness under physical hardware constraints have yet to be fully explored. Second, although our action chunk-level optimization improves short-term coherence, the architecture currently lacks a dedicated world model for structured long-horizon reasoning and environmental transition prediction.

Moving forward, we plan to deploy OmniVLA-RL on physical robotic platforms to evaluate its real-world reliability and adaptation capabilities. To address the planning bottleneck, we aim to integrate a generative world model, enabling the agent to perform ``imagination-based'' planning. Furthermore, we will extend our framework to handle more unstructured, open-world manipulation tasks involving diverse objects and multi-modal sensory feedback.

\bibliographystyle{apalike}
\bibliography{references}

\end{document}